\documentclass{Interspeech2024}
\usepackage{svg}
\usepackage{subcaption}
\usepackage{xurl}

\interspeechcameraready

\title{MaViLS, a Benchmark Dataset for Video-to-Slide Alignment, Assessing Baseline Accuracy with a Multimodal Alignment Algorithm Leveraging Speech, OCR, and Visual Features\vspace{-0.65cm}}

\name[affiliation={1,2}]{Katharina}{Anderer}
\name[affiliation={3}]{Andreas}{Reich}
\name[affiliation={1,3}]{Matthias}{Wölfel}

\address{$^1$Karlsruhe University of Applied Sciences, Germany\\
  $^2$Karlsruhe Institute of Technology, Germany\\
  $^3$University of Hohenheim, Germany}
\email{katharina.anderer@kit.edu, andreas.reich@uni-hohenheim.de, matthias.woelfel@h-ka.de}


\begin{document}

\maketitle

\begin{abstract}

This paper presents a benchmark dataset for aligning lecture videos with corresponding slides and introduces a novel multimodal algorithm leveraging features from speech, text, and images. It achieves an average accuracy of 0.82 in comparison to SIFT (0.56) while being approximately 11 times faster. Using dynamic programming the algorithm tries to determine the optimal slide sequence. The results show that penalizing slide transitions increases accuracy. Features obtained via optical character recognition (OCR) contribute the most to a high matching accuracy, followed by image features. The findings highlight that audio transcripts alone provide valuable information for alignment and are beneficial if OCR data is lacking. Variations in matching accuracy across different lectures highlight the challenges associated with video quality and lecture style. The novel multimodal algorithm demonstrates robustness to some of these challenges, underscoring the potential of the approach.
\end{abstract}

\section{Introduction}
Latest since the COVID-19 pandemic, online lectures have become an integral part of academic settings. The shift towards virtual education brings forth both challenges and opportunities to ensure that educational content is not only accessible but also engaging and tailored to meet diverse learning preferences \cite{ahel2020opportunities}.
Integrating video lectures with corresponding slide presentations enhances the learning experience by combining auditory and visual elements. This multimodal approach is particularly beneficial when learners have impaired auditory or visual processing, reflecting findings by \cite{el2020presenting, himmelsbach2015enabling} that emphasize the importance of not relying solely on a single sensory modality in accessible software and web design.
Moreover, content revision and self-paced study can be enhanced. A lecture tutoring system called PET \cite{wolfel2021towards,bdcc8010002} aligns user questions to corresponding slides and navigates the user to the best-fitting slide by calculating the similarity between the slide text and the user question. If lecture slides are aligned with the lecture's speech as well, this could enhance navigation through slides to a large degree. The synchronization of audio and slides can also enhance chatbot functionality in general, improving their ability to link user questions with the most relevant content. 
Another application could be the improvement of alternative text descriptions for visuals. Visuals on slides often lack alternative text descriptions \cite{lundgard2021accessible} leading to a lack of information for visually impaired people. If a lecturer is talking about the visuals on a particular slide, the aligned audio transcript can be used to automatically generate alternative text about these visuals.

The task of matching video frames to lecture slides can be quite challenging under circumstances where e.g. lectures include demonstrations, external videos, or web pages not present in the slides, or where instructors navigate slides non-linearly in response to student questions. Additional challenges arise from poor video or audio quality, which can hinder audio transcription or OCR accuracy, as well as from suboptimal recording angles or videos focusing on the instructor rather than the slides.

Previous research has employed methods like scale-invariant feature transform (SIFT) \cite{fan2006matching, fan2011robust, jeong2012accurate, wang2009robust} for slide alignment, an algorithm first proposed by \cite{lowe2004distinctive}. Others have used global pixel differences or color histograms \cite{eruvaram2020experimental}.  
A tool called \textit{Talkminer}, introduced by \cite{adcock2010talkminer}, presents a search engine for webcasts that uses OCR to extract keywords, enabling users to locate specific video sections using a keyword search \cite{adcock2010talkminer}. A study by \cite{jones2002automated} already aligns electronic slides with the audio transcript, but with a focus solely on OCR features and assuming a strictly linear sequence of slides, not allowing for possible backward jumps \cite{jones2002automated}. \textit{Slidecho}, a tool introduced by \cite{peng2021slidecho} extracts slides from lecture videos using OCR to identify the most text-rich bounding boxes and align them with speech. Limitations of this tool are that it disregards visual features and does not match video frames with high-quality PDF slides. High-resolution PDF slides can encapsulate text and graphical elements more effectively and may even accommodate alternative text descriptions for images or tables, enhancing accessibility for individuals with visual impairment, when saved in an accessible format such as PDF/UA. The presence of PDF slides can therefore play a vital role in preserving information when aligning them with video frames.
Our approach combines multiple features, namely OCR, image features, and audio features, making it more robust than algorithms that consider either text or image features. It matches video frames to PDF slides to preserve as much information as possible. Additionally, it presents a more efficient method than SIFT which is rather time-consuming.  

We publish a broad benchmark dataset, including 20 labeled lectures of various study fields, varying lengths, and different lecturers such that various presenting styles are included. 
This dataset, which is published together with our novel alignment algorithm on Github\footnote{\url{https://github.com/andererka/MaViLS}}, provides a diverse open-source dataset concerning domains, video and speech quality and lecturers and enables a structured analysis of different alignment algorithms.

\section{MaViLS Dataset}

The dataset MaViLS, an acronym for 'Matching Videos to Lecture Slides', includes ground truth files for 20 lectures from various fields. To create a diverse dataset, we selected lectures from the medical, engineering, and natural sciences, including a lecture on computer vision, psychology, reinforcement learning, climate science, numerics, and game design.
Most of the lectures are taken from the MIT OpenCourseWare\footnote{\url{https://ocw.mit.edu/}}, a broad selection of lectures with corresponding material from the Massachusetts Institute of Technology. Furthermore, two lectures by the University of Tuebingen, Germany, and one by DeepMind are selected. 

MaViLS includes the audio transcripts of the lectures, the videos, the corresponding PDF slides, and the ground truth files that match slides to video frames and spoken sentences. The audio transcription is done by faster-whisper\footnote{\url{https://github.com/SYSTRAN/faster-whisper}} with 1550 million parameters, an efficient reimplementation of OpenAI's speech recognition tool Whisper \cite{radford2023robust}. For the ground truth files, each sentence of a lecturer is mapped to a slide manually by human raters and has a video timestamp. If either no slide is captured or the slide is not uniquely identifiable, '$-1$' is returned as the slide label. 

Designed to test the robustness of matching algorithms, MaViLS intentionally features lectures with challenges such as low-quality videos, frequent perspective shifts, and slides with minor differences. The \textit{video image} and  \textit{audio quality} are assessed on a $7$-point Likert scale, where $1$ indicates inferior quality and $7$ very good quality. 
Further, a \textit{volatility score} is calculated that describes how often the lecturer jumps back and forth in the slides according to the total number of slides. It is calculated as the ratio between the count of slide number changes and the total amount of slides. A number close to $1$ indicates that the number of slide changes was close to the amount of slides, while a high number (e.g. above 1.5) indicates a relatively high volatility.
Additionally, the \textit{no slide/slide ratio} is indicating the prevalence of slides shown for each video frame in comparison to video frames without slides. The ratio varies largely across lectures, from zero where every frame is showing a slide, to a ratio of ten, indicating that a frame without a slide is ten times more likely compared to a frame capturing a slide. 

MaViLS encompasses over 22 hours of video content, amounting to 12,830 distinct video segments with accompanying audio transcripts and slide labels.

\section{Methods}
\label{section:methods}
We propose a novel MaViLS algorithm that utilizes a combination of features, including text utilizing OCR, visual information, and audio transcripts to accurately link each video frame to the most similar lecture slide.

\subsection{Features}
To extract text from the video frames and lecture slides, we employ the open-source framework Tesseract \cite{smith2009adapting}. Subsequently, the sentence transformer model '\textit{distiluse-base-multilingual-cased}' available on Hugging Face\footnote{\url{https://huggingface.co/sentence-transformers/ distiluse-base-multilingual-cased}} is used to create embeddings of the extracted texts. Cosine similarity is then computed between the text embeddings to produce a similarity matrix for each slide and video frame pair. This constructed similarity matrix encapsulates the text features.

For the audio transcript of each video frame and the extracted text from lecture slides, an analogous procedure is employed. The sentence transformer model converts the texts to embeddings. The cosine similarities between these embeddings provide a second similarity matrix, representing the speech features.

For visual feature extraction, we choose the transformer model '\textit{MBZUAI/swiftformer-xs}'\footnote{\url{https://huggingface.co/MBZUAI/swiftformer-xs}} for its high performance regarding efficiency and accuracy\footnote{\url{https://doi.org/10.48550/arXiv.2303.15446}}. To derive the image feature embeddings, both video frames and slides are input into the model. The resultant last hidden states form the basis of our features. Computing cosine similarities between these embeddings gives a third matrix, representing the image-based features.

These three similarity matrices serve in the next step as the foundation for the subsequent determination of the optimal sequence of slides accompanying a lecture video.

\subsection{Dynamic programming optimization}

Our approach uses dynamic programming to systematically choose the best sequence of slides to accompany a video. We construct a decision matrix, hereafter referred to as the $D$ matrix, where each entry $D_{i, j}$ represents the highest cumulative similarity score when matching video frame $i$ with slide $j$.

Following, $i$ designates the index of a video frame and \textit{j} designates the index of the current lecture slide. A third index $k$ is used for the optimization formula below corresponding to the lecture slide index with the potentially highest cumulative score. The similarity score, denoted by $S_{i, j}$, is a measure of how well video frame \textit{i} matches with slide $j$. $S_{i, j} \in [-1, 1]$, where 1 defines the highest possible similarity.

The $D$ matrix maintains a record of the optimal cumulative scores. Each entry $D_{i, j}$ is derived by locating the best score from the previous step, denoted as $D_{i-1,j}(k)$, and adding the current similarity score $S_{i, j}$. Simultaneously, a penalty, $p_j^{\text{jump}}$, is imposed to discourage slide transitions that are likely coming from noise, while a reward for linear succession of slides, $p_i^{\text{linear}}$, is also factored in. This results in the following formula for a particular entry for $D$:

\begin{align}
\label{equ:DP}
D_{i, j} &= \max_{k} \big( D_{i-1, j}(k) - p_j^{\text{jump}}(k) - p_i^{\text{linear}} \big) + S_{i, j}
\end{align}

Penalty $p_j^{\text{jump}}(k)$ imposes a penalty proportional to the magnitude of the jump. The penalty is higher for backward transitions ($k~<~j$) as compared to forward ones ($k~>~j$):
\begin{equation}
p_j^{\text{jump}}(k) = 
\begin{cases} 
2 \cdot |k-j| \cdot \lambda^{\text{jump}} & k < j \\
0 & k = j \\
|k-j| \cdot \lambda^{\text{jump}} & k > j,\\ 
\end{cases}
\end{equation}
where $\lambda^{\text{jump}}$ is a small constant for weighting the penalty.

$p_i^{\text{linear}}$ presents an incentive to follow a sequential order considering the expected slide index at a given frame. The expected slide index $e_i$ is deduced from the ratio of the total slide number $m$ to the total frame number $n$ and is calculated as:
\begin{equation}
  e_i = 1 + \frac{m}{n-1}\cdot i  
\end{equation}

$p_i^{\text{linear}}$ is then defined as the deviation between the expected frame index and the actual index, multiplied by a small constant $\lambda^{\text{linear}}$:
\begin{equation}
  p_i^{\text{linear}} = |e_i - i| \cdot \lambda^{\text{linear}}  
\end{equation}

In the results section different values for $\lambda^{\text{jump}}$ ($0$, $0.1$, $0.15$, $0.2$, $0.25$) and for $\lambda^{\text{linear}}$ ($0$, $10^{-4}$, $10^{-3}$) are compared. 

\subsection{Combination techniques}

The final step integrates the text-, image feature-, and audioscript-based similarity matrices into a combined matrix, which then feeds into the dynamic programming optimization process. As first approach, hereinafter referred to as 'mean' combination, the average value for each entry $i, j$ is calculated, maintaining matrix dimensions for the combined similarity matrix. Similarly, as a second approach, the maximum value for each entry $i, j$  is calculated ('max' combination).

A third approach adopts a weighted sum of the three matrices. To identify the optimal weights \textit{$w_a, w_b, w_c$}, gradient descent is performed per lecture, aiming to maximize the cumulative similarity score.

The combined similarity matrix $S$ is calculated using the single feature matrices $A, B, C$ in the following way:

\begin{equation}
\label{equ_S_weighted}
S = w_a \cdot A + w_b \cdot B + w_c \cdot C
\end{equation}
Gradient descent is started with all initial weights set equally to $w_a = w_b = w_c = 1/3$.

The objective is to maximize the $D$ score per lecture (Eq.~\ref{equ:DP}) by finding the optimal weights, using $S$ as defined in Eq.~\ref{equ_S_weighted}. Gradient descent with an integrated Adam optimizer\footnote{\url{https://doi.org/10.485
50/arXiv.1412.6980}} is deployed, using a learning rate of 0.001 and a total of 50 iterations to keep computational costs low.







\section{Results}
To assess the accuracy of different algorithms, we compute recall, precision and F1 scores. The task is defined as a binary classification and for each video frame it is determined whether the slide number was correctly aligned or not, ignoring labels of '$-1$' where no slide was captured. The variation between F1, recall, and precision scores is notably minimal across our findings. For better readability, we only discuss F1 scores subsequently.
Table~\ref{tab:features} summarizes the F1 scores for every lecture.

\begin{figure}[t]
\label{fig:video_quality}
  \centering
  \begin{subfigure}{0.49\linewidth}
    \includegraphics[width=\linewidth]{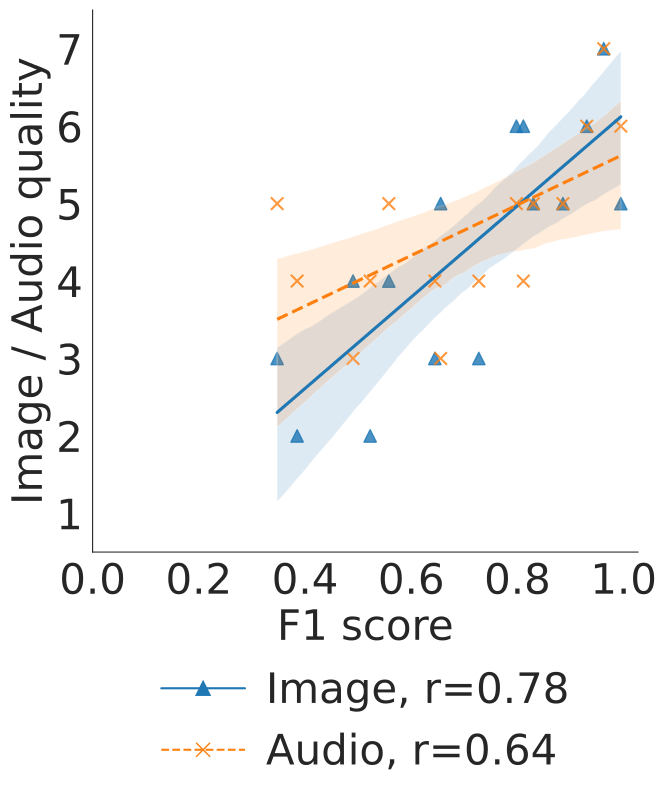}
    \caption{Text based algorithm}
    \label{fig:ocr_quality}
  \end{subfigure}
  \begin{subfigure}{0.49\linewidth}
    \includegraphics[width=\linewidth]{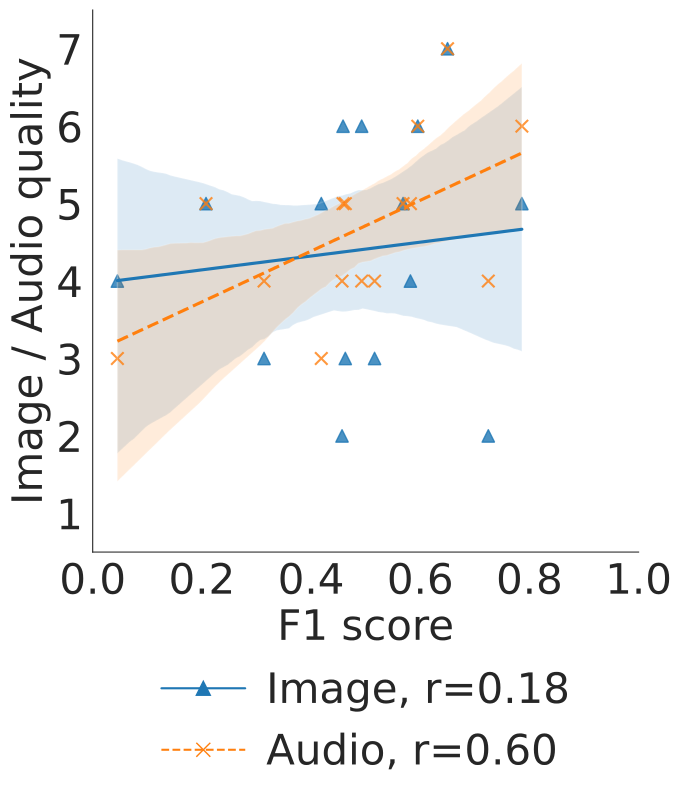}
    \caption{Audio based algorithm}
    \label{fig:audio_quality}
  \end{subfigure}
  \caption{Correlation between F1 score and video quality (7-point Likert). Image quality is shown in blue (solid line, triangle symbols), audio quality in orange (dashed line, cross symbols). Regression lines and 95\% confidence intervals (CI) are shown.}
  \label{fig:video_quality}
  \vspace{-3mm}
\end{figure}

\subsection{Features}
\label{subsec:features}
Text embeddings achieve the highest average score with 0.76 accuracy, succeeded by image embeddings (0.64). Audio embeddings also prove valuable, achieving an average of 0.53. For 'Product Design' audio embeddings even demonstrate superior outcomes over the other embeddings. 

\begin{table}[t]
  \caption{F1 score for single feature MaViLS and SIFT baseline. }
  \label{tab:features}
  \centering
\begin{tabular}{lrrr|r}
\toprule
                     \textbf{Lecture name} &  \textbf{Text} &  \textbf{Audio} &  \textbf{Image} &  \textbf{SIFT} \\
\midrule
        Deep learning &     0.99 &       0.79 &        0.93 &  0.49 \\
          Short range &     0.89 &       0.57 &        0.73 &  0.62 \\
             Numerics &     0.80 &       0.46 &        0.50 &  0.82 \\
        Reinforcement &     0.64 &       0.31 &        0.00 &  0.44 \\
      Computer Vision &     0.96 &       0.65 &        0.99 &  0.93 \\
     Climate \& Cities &     0.81 &       0.49 &        0.82 &  0.51 \\
      Decarbonization &     0.73 &       0.52 &        0.89 &  0.50 \\
       Cryptocurrency &     0.83 &       0.21 &        0.70 &  0.66 \\
       Solar resource &     0.35 &       0.46 &        0.53 &  0.43 \\
           Psychology &     0.56 &       0.58 &        0.76 &  0.31 \\
        Productdesign &     0.39 &       0.72 &        0.41 &  0.78 \\
     Image processing &     0.99 &       0.53 &        0.06 &  0.25 \\
      Sensory systems &     0.52 &       0.46 &        0.58 &  0.17 \\
        ML for health &     0.93 &       0.57 &        0.95 &  0.44 \\
     Climate policies &     0.91 &       0.79 &        0.64 &  0.83 \\
Computation theory &     0.86 &       0.56 &        0.83 &  0.58 \\
              Physics &     0.93 &       0.60 &        0.92 &  0.60 \\
            Phonetics &     0.49 &       0.05 &        0.81 &  0.53 \\
        Team dynamics &     0.95 &       0.89 &        0.25 &  0.92 \\
   Cognitive robotics &     0.66 &       0.42 &        0.55 &  0.44 \\
   \midrule
                    \textbf{Average } &     \textbf{0.76} &       \textbf{0.53} &        \textbf{0.64} &        \textbf{0.56}  \\

\bottomrule
\end{tabular}
\vspace{-3mm}
\end{table}

These observations suggest a contextual effectiveness of each embedding; no single embedding method uniformly outperforms the others across lectures. For instance, in scenarios where slides are devoid of text, text embeddings are naturally at a disadvantage. Further, audio embeddings can give valuable information in cases where the image quality of the video frame is inferior.
Correlating audio and image quality with the accuracy of text and audio embeddings confirms this argument. Text feature accuracy correlates strongly with the image quality of the video ($r= 0.78$), whereas the audio-based algorithm is not affected by the image quality ($r= 0.18$), but is affected by the audio quality of the video ($r= 0.60$) as shown in Fig.~\ref{fig:video_quality}.

$\lambda^{\text{jump}}=0.1$ is used for comparison, as this is the best-performing penalty value on average (see Sec.~\ref{subsection:jum_penalty}).


\subsection{Combination techniques}

An analysis of the different methods for combining the text, audio, and image similarity matrices does not reveal any superior approach. 
On average, all combination techniques (max, mean, weighted sum) achieve an accuracy of 0.82, significantly higher than the accuracies of the single feature algorithms reported in Table~\ref{tab:features}. For a baseline, SIFT matching is computed achieving an average accuracy of 0.56 (see Table~\ref{tab:features}). The python cv2 package\footnote{\url{https://docs.opencv.org/}} with the functions \textbf{SIFT\_create} and \textbf{BFMatcher} are used to compute local features and to match them between PDF slides and video frames. 
$\lambda^{\text{jump}}=0.1$ is used for comparing the combination techniques here.


\subsection{Penalties}
\label{subsection:jum_penalty}

For the linear penalty term $p_i^{\text{linear}}$, setting $\lambda^{\text{linear}}=0$ yields the most favorable outcome, with higher penalty values leading to a decline in performance across most cases. However, for two lectures, the penalty $\lambda^{\text{linear}}=10^{-3}$ results in slight enhancements, indicating that a penalty for rewarding a linear slide order might be beneficial under some circumstances. 

\begin{figure}[t]
  \centering
  \begin{subfigure}{0.49\linewidth}
    \includegraphics[width=\linewidth]{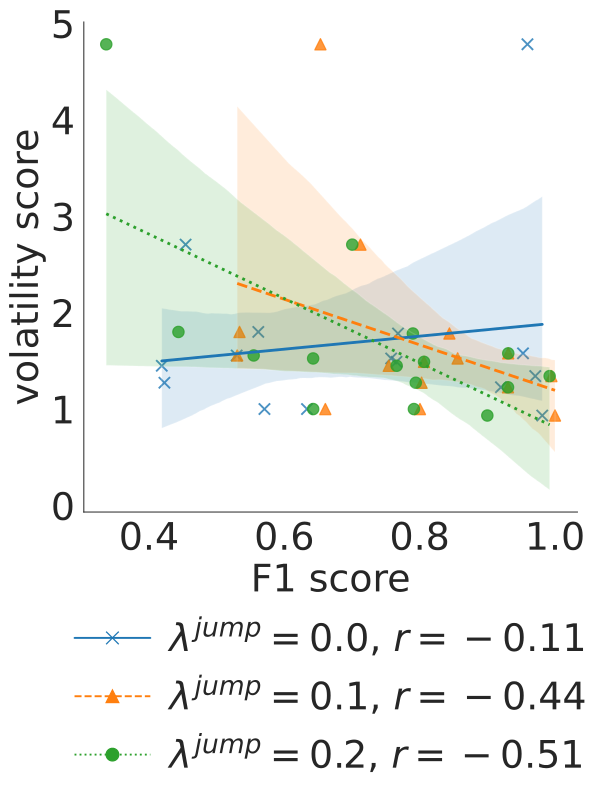}
  \end{subfigure}
  \begin{subfigure}{0.49\linewidth}
    \includegraphics[width=\linewidth]{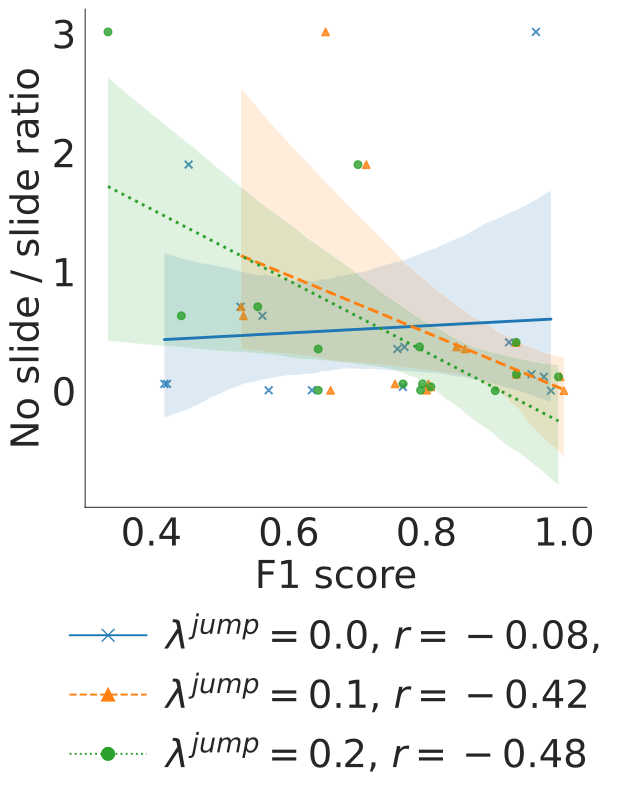}
  \end{subfigure}
  \caption{Left: Correlation between F1 and volatility scores. Right: Correlation between F1 and 'no slide / slide ratio' scores. Regression lines and 95\% CI are plotted. Blue (cross, solid) relates to $\lambda^{\text{jump}}=0$, orange (triangle, dashed) to $\lambda^{\text{jump}}=0.1$ and green (point, dotted) to $\lambda^{\text{jump}}=0.2$}
  \label{fig:volatility_no_slide}
  \vspace{-3mm}
\end{figure}

For $p_j^{\text{jump}}(k)$, setting $\lambda^{\text{jump}}=0.1$ achieves the highest accuracy, yielding an average score of $0.82$, compared to the average score of $0.76$ for $\lambda^{\text{jump}}=0$ as reported in Table~\ref{tab:jump_penalty}. Setting $\lambda^{\text{jump}}=0.15$ or higher decreases performance again for the majority of lectures, although it does show slightly enhanced performance for 'Cognitive Robotics' and 'Reinforcement Learning', indicating that these lectures profit from a higher penalty. 

\begin{table}[t]
  \caption{F1 score for different $\lambda^{\text{jump}}$ for multimodal MaViLS.}
  \label{tab:jump_penalty}
  \centering
\begin{tabular}{lrrrrr}
\toprule
 \textbf{Lecture name} & \multicolumn{5}{c}{\textbf{$\mathbf{\lambda^{\text{jump}}}$}} \\ \cline{2-6}
     &  \textbf{0.0} &  \textbf{0.1} & \textbf{0.15} & \textbf{0.2} & \textbf{0.25}  \\
\midrule
        Deep learning &         0.97 &         0.99 &          0.99 &         0.99 &          0.99 \\
          Short range &         0.42 &         0.80 &          0.80 &         0.79 &          0.75 \\
             Numerics &         0.77 &         0.81 &          0.80 &         0.81 &          0.79 \\
        Reinforcement &         0.42 &         0.75 &          0.78 &         0.77 &          0.77 \\
      Computer Vision &         0.98 &         1.00 &          0.99 &         0.90 &          0.90 \\
     Climate \& Cities &         0.76 &         0.86 &          0.86 &         0.64 &          0.81 \\
      Decarbonization &         0.95 &         0.93 &          0.93 &         0.93 &          0.89 \\
       Cryptocurrency &         0.77 &         0.84 &          0.82 &         0.79 &          0.79 \\
       Solar resource &         0.56 &         0.53 &          0.53 &         0.44 &          0.35 \\
           Psychology &         0.63 &         0.80 &          0.79 &         0.79 &          0.80 \\
        Productdesign &         0.45 &         0.71 &          0.71 &         0.70 &          0.64 \\
     Image processing &         0.91 &         0.97 &          0.97 &         0.97 &          0.96 \\
      Sensory systems &         0.53 &         0.53 &          0.53 &         0.55 &          0.55 \\
        ML for health &         0.92 &         0.95 &          0.95 &         0.94 &          0.94 \\
     Climate policies &         0.93 &         0.93 &          0.95 &         0.95 &          0.94 \\
Computation theory &         0.86 &         0.84 &          0.84 &         0.87 &          0.87 \\
              Physics &         0.92 &         0.93 &          0.93 &         0.93 &          0.93 \\
            Phonetics &         0.96 &         0.65 &          0.37 &         0.34 &          0.34 \\
        Team dynamics &         1.00 &         0.96 &          0.96 &         0.96 &          0.64 \\
   Cognitive robotics &         0.57 &         0.66 &          0.68 &         0.64 &          0.62 \\
 \bottomrule                 \textbf{Average} &         \textbf{0.76} &         \textbf{0.82} &          \textbf{0.81} &         \textbf{0.79} &         \textbf{0.76} \\

\bottomrule
\end{tabular}
\vspace{-3mm}
\end{table}

A higher jump penalty leads to a stronger negative correlation between F1 score and \textit{volatility score} as depicted in Fig.~\ref{fig:volatility_no_slide}. This is not surprising as lectures with a high volatility, like for instance 'Phonetics' that has a volatility score of 4.78, perform worse if a high jump penalty is introduced which punishes jumping back and forth. Equally, a higher jump penalty leads to a stronger negative correlation between F1 score and the \textit{no slide/slide ratio} which is depicted on the right side of Fig.~\ref{fig:volatility_no_slide}.  

In summary, introducing a jump penalty is beneficial on average, but might be counterproductive for lectures with high volatility and a high occurrence of 'no slide' views.

\subsection{Runtime} 
Given the transcribed audio, calculating features and alignment for only audio features takes around 45 seconds on our local CPU (i7, 3.0 GHz), for image features it takes around 64 seconds and for text features it takes three minutes on average.
The combined alignment using either 'max' or 'mean' combination takes 3.5 minutes in sum, while the weighted sum approach takes around 4.6 minutes. SIFT alignment takes 39.5 minutes on average, being the most time-consuming. Therefore, the MaViLS algorithm is approximately 11 times faster.



 
\section{Discussion}

This paper analyses the effectiveness of single-feature embeddings versus combined-feature embeddings, demonstrating that multi-feature composites improve the alignment of lecture slides with the corresponding audio transcripts or video frames. Potentially, adding more algorithms or embeddings to the mix could further increase the results. However, one must balance between high multimodality and time efficiency.

No notable performance disparities are observed among the three combination techniques. The weighted sum method is less effective when lectures vary between parts where text excels and segments where audio is more informative, since it uniformly weights across the entire lecture rather than adjusting to individual video frames. As the approach is additionally slower than the other techniques, its utility seems limited.

Further, the implementation of a jump penalty that discourages frequent switching between slide numbers is explored. While incorporating such a penalty bolsters accuracy on average, it is counterproductive in lectures with frequent slide switches. Future research could investigate methods for detecting volatility. Also, alternative (e.g. non-linear) jump penalty calculations might be beneficial. 

Another key insight of this paper is that the audio transcript features are more resilient to suboptimal quality of the recordings than text-based features, offering valuable information for alignment even with no text captured by the video frame. This underscores the power of a multimodal approach.

The creation of rich embedding spaces presents opportunities for digital academic settings. Enhanced matching capabilities can facilitate students' inquiries, linking questions to relevant lecture content more efficiently. This can be particularly beneficial for visually impaired individuals by synchronizing spoken content with visual materials.

MaViLS is limited to English language lectures and might lack comprehensive field representation. Domains such as mathematics are underrepresented, as lectures from these areas seldom come with accompanying PDF slides. 
Further, some images are shown in the video, but not on the PDF slides due to copyright restrictions, which may impact our results.

In addition, embeddings are calculated on a sentence level only. Flexible text blocks can potentially improve alignment, presenting an interesting direction for future research.

\section{Acknowledgements}
Andreas Reich is supported by 'Stiftung Innovation in der Hochschullehre' (project number FBM2020-EA-1670-01800). Katharina Anderer is supported by the Ph.D. scholarship in the Cooperative Doctoral Program Accessibility through AI-based Assistive Technology (kooperatives Promotionskolleg Barrierefreiheit durch KI-Basierte Assistive Technologien (KATE)) provided by the Ministry of Science, Research and Arts Baden-Württemberg (MWK).

\bibliographystyle{IEEEtran}
\bibliography{mybib}

\begin{thebibliography}{10}
\providecommand{\url}[1]{#1}
\csname url@samestyle\endcsname
\providecommand{\newblock}{\relax}
\providecommand{\bibinfo}[2]{#2}
\providecommand{\BIBentrySTDinterwordspacing}{\spaceskip=0pt\relax}
\providecommand{\BIBentryALTinterwordstretchfactor}{4}
\providecommand{\BIBentryALTinterwordspacing}{\spaceskip=\fontdimen2\font plus
\BIBentryALTinterwordstretchfactor\fontdimen3\font minus \fontdimen4\font\relax}
\providecommand{\BIBforeignlanguage}[2]{{%
\expandafter\ifx\csname l@#1\endcsname\relax
\typeout{** WARNING: IEEEtran.bst: No hyphenation pattern has been}%
\typeout{** loaded for the language `#1'. Using the pattern for}%
\typeout{** the default language instead.}%
\else
\language=\csname l@#1\endcsname
\fi
#2}}
\providecommand{\BIBdecl}{\relax}
\BIBdecl

\bibitem{ahel2020opportunities}
O.~Ahel and K.~Lingenau, ``Opportunities and challenges of digitalization to improve access to education for sustainable development in higher education,'' in \emph{Universities as Living Labs for Sustainable Development}.\hskip 1em plus 0.5em minus 0.4em\relax Cham, Switzerland: Springer, 2020, pp. 341--356.

\bibitem{el2020presenting}
Y.~El-Glaly, W.~Shi, S.~Malachowsky, Q.~Yu, and D.~E. Krutz, ``Presenting and evaluating the impact of experiential learning in computing accessibility education,'' in \emph{Proceedings of the ACM/IEEE 42nd International Conference on Software Engineering: Software Engineering Education and Training}, Seoul, Republic of Korea, May 2020, pp. 49--60.

\bibitem{himmelsbach2015enabling}
J.~Himmelsbach, M.~Garschall, S.~Egger, S.~Steffek, and M.~Tscheligi, ``Enabling accessibility through multimodality? interaction modality choices of older adults,'' in \emph{Proceedings of the 14th International Conference on Mobile and Ubiquitous Multimedia}, Nov. 2015, pp. 195--199.

\bibitem{wolfel2021towards}
M.~W{\"o}lfel, ``Towards the automatic generation of pedagogical conversational agents from lecture slides,'' in \emph{Multimedia Technology and Enhanced Learning: Third EAI International Conference, ICMTEL 2021, Virtual Event, April 8--9, 2021, Proceedings, Part II 3}.\hskip 1em plus 0.5em minus 0.4em\relax Springer, 2021, pp. 216--229.

\bibitem{bdcc8010002}
\BIBentryALTinterwordspacing
M.~Wölfel, M.~B. Shirzad, A.~Reich, and K.~Anderer, ``Knowledge-based and generative-ai-driven pedagogical conversational agents: A comparative study of grice’s cooperative principles and trust,'' \emph{Big Data and Cognitive Computing}, vol.~8, no.~1, 2024. [Online]. Available: \url{https://www.mdpi.com/2504-2289/8/1/2}
\BIBentrySTDinterwordspacing

\bibitem{lundgard2021accessible}
A.~Lundgard and A.~Satyanarayan, ``Accessible visualization via natural language descriptions: A four-level model of semantic content,'' \emph{IEEE transactions on visualization and computer graphics}, vol.~28, no.~1, pp. 1073--1083, Sep. 2021.

\bibitem{fan2006matching}
Q.~Fan, K.~Barnard, A.~Amir, A.~Efrat, and M.~Lin, ``Matching slides to presentation videos using sift and scene background matching,'' in \emph{Proceedings of the 8th ACM international workshop on Multimedia information retrieval}, Santa Barbara, California, USA, Oct. 2006, pp. 239--248.

\bibitem{fan2011robust}
Q.~Fan, K.~Barnard, A.~Amir, and A.~Efrat, ``Robust spatiotemporal matching of electronic slides to presentation videos,'' \emph{IEEE transactions on image processing}, vol.~20, no.~8, pp. 2315--2328, Jan. 2011.

\bibitem{jeong2012accurate}
H.~J. Jeong, T.-E. Kim, and M.~H. Kim, ``An accurate lecture video segmentation method by using sift and adaptive threshold,'' in \emph{Proceedings of the 10th International Conference on Advances in Mobile Computing \& Multimedia}, Bali, Indonesia, Dec. 2012, pp. 285--288.

\bibitem{wang2009robust}
X.~Wang and M.~Kankanhalli, ``Robust alignment of presentation videos with slides,'' in \emph{Advances in Multimedia Information Processing-PCM 2009: 10th Pacific Rim Conference on Multimedia}.\hskip 1em plus 0.5em minus 0.4em\relax Bangkok, Thailand: Springer, Dec. 2009, pp. 311--322.

\bibitem{lowe2004distinctive}
D.~G. Lowe, ``Distinctive image features from scale-invariant keypoints,'' \emph{International journal of computer vision}, vol.~60, pp. 91--110, 2004.

\bibitem{eruvaram2020experimental}
P.~Eruvaram, K.~Ramani, and C.~S. Bindu, ``An experimental comparative study on slide change detection in lecture videos,'' \emph{International Journal of Information Technology}, vol.~12, no.~2, pp. 429--436, 2020.

\bibitem{adcock2010talkminer}
J.~Adcock, M.~Cooper, L.~Denoue, H.~Pirsiavash, and L.~A. Rowe, ``Talkminer: a lecture webcast search engine,'' in \emph{Proceedings of the 18th ACM international conference on Multimedia}, Firenze, Italy, Oct. 2010, pp. 241--250.

\bibitem{jones2002automated}
G.~J.~F. Jones and R.~J. Edens, ``Automated alignment and annotation of audio-visual presentations,'' in \emph{Research and Advanced Technology for Digital Libraries}, M.~Agosti and C.~Thanos, Eds.\hskip 1em plus 0.5em minus 0.4em\relax Berlin, Heidelberg: Springer Berlin Heidelberg, 2002, pp. 276--291.

\bibitem{peng2021slidecho}
Y.-H. Peng, J.~P. Bigham, and A.~Pavel, ``Slidecho: Flexible non-visual exploration of presentation videos,'' in \emph{Proceedings of the 23rd International ACM SIGACCESS Conference on Computers and Accessibility}, Virtual Event, USA, Oct. 2021, pp. 1--12.

\bibitem{radford2023robust}
\BIBentryALTinterwordspacing
A.~Radford, J.~W. Kim, T.~Xu, G.~Brockman, C.~McLeavey, and I.~Sutskever, ``Robust speech recognition via large-scale weak supervision,'' in \emph{International Conference on Machine Learning}.\hskip 1em plus 0.5em minus 0.4em\relax Honolulu, Hawaii, USA: PMLR, 2023, pp. 28\,492--28\,518. [Online]. Available: \url{https://proceedings.mlr.press/v202/radford23a.html}
\BIBentrySTDinterwordspacing

\bibitem{smith2009adapting}
R.~Smith, D.~Antonova, and D.-S. Lee, ``Adapting the tesseract open source ocr engine for multilingual ocr,'' in \emph{Proceedings of the International Workshop on Multilingual OCR}, 2009, pp. 1--8.

\end{thebibliography}

\end{document}